\documentclass[runningheads]{llncs}
\usepackage[T1]{fontenc}
\usepackage[table,xcdraw]{xcolor}
\usepackage{graphicx}
\usepackage{booktabs}
\usepackage{hyperref}
\usepackage{amsmath}
\usepackage{amssymb}
\usepackage{multirow}
\usepackage{enumitem}
\usepackage{subcaption}
\usepackage{float}      
\usepackage{setspace}   
\usepackage[font=small]{caption}

\usepackage{pifont}
\usepackage[misc]{ifsym}
\newcommand{\corr}{(\Letter)}
\usepackage{mwe}
\begin{document}


\title{ETT-CKGE: Efficient Task-driven Tokens for Continual Knowledge Graph Embedding}

\titlerunning{Efficient Task-driven Tokens for Continual Knowledge Graph Embedding}

\author{Lijing Zhu\inst{1}
\and Qizhen Lan\inst{6}
\and Qing Tian\inst{6}
\and Wenbo Sun \inst{3}
\and Li Yang \inst{4}
\and Lu Xia \inst{7}
\and Yixin Xie \inst{5}
\and Xi Xiao \inst{6}
\and Tiehang Duan \inst{2}
\and Cui Tao \inst{2}
\and \corr{Shuteng Niu}\inst{1,2}
}

\authorrunning{L. Zhu at al}

\institute{Bowling Green State University, Bowling Green OH 43403, USA.
\and
Mayo Clinic, Jacksonville, FL 32066, USA 
\and
University of Michigan, Ann Arbor, MI 48109, USA
\and
University of North Carolina, Charlotte, NC 28223, USA
\and
Kennesaw State University, Kennesaw, GA 30144, USA
\and
University of Alabama, Birmingham, AL 35294, USA
\and 
Michigan State University, East Lansing, MI 48824, USA
}

\maketitle              
\sloppy

\begin{abstract}
Continual Knowledge Graph Embedding (CKGE) seeks to integrate new knowledge while preserving past information. However, existing methods struggle with efficiency and scalability due to two key limitations: (1) suboptimal knowledge preservation between snapshots caused by manually designed node/relation importance scores that ignore graph dependencies relevant to the downstream task, and (2) computationally expensive graph traversal for node/relation importance calculation, leading to slow training and high memory overhead. To address these limitations, we introduce \textbf{ETT-CKGE} (\textbf{E}fficient, \textbf{T}ask-driven, \textbf{T}okens for \textbf{C}ontinual \textbf{K}nowledge \textbf{G}raph \textbf{E}mbedding), a novel task-guided CKGE method that leverages efficient task-driven tokens for efficient and effective knowledge transfer between snapshots. Our method introduces a set of learnable tokens that directly capture task-relevant signals, eliminating the need for explicit node scoring or traversal. These tokens serve as consistent and reusable guidance across snapshots, enabling efficient token-masked embedding alignment between snapshots. Importantly, knowledge transfer is achieved through simple matrix operations, significantly reducing training time and memory usage. Extensive experiments across six benchmark datasets demonstrate that ETT-CKGE consistently achieves superior or competitive predictive performance, while substantially improving training efficiency and scalability compared to state-of-the-art CKGE methods. The code is available at \href{https://github.com/lijingzhu1/ETT-CKGE/tree/main}{Github}.

\end{abstract}

\section{Introduction}

Knowledge graph embedding (KGE) aims to project nodes and relations into a continuous vector space to support downstream applications~\cite{li2025improving,ju2024comprehensive,10825929,xiao2024hgtdp} such as node classification, knowledge graphs (KGs) completion, and graph classification. While traditional KGE methods primarily focus on static KGs~\cite{wang2017knowledge}, real-world KGs are inherently dynamic, continuously evolving with emerging nodes, relations, and facts. In such settings, retraining KGE models from scratch becomes computationally expensive. To address this challenge, Continual Knowledge Graph Embedding (CKGE)~\cite{cui2023lifelong} has been proposed as a practical paradigm that incrementally updates node and relation representations over a sequence of knowledge graph (KG) snapshots while mitigating catastrophic forgetting of previously learned knowledge. 

Generally, previous research has explored approaches such as parameter isolation \cite{rusu2016progressive,lomonaco2017core50}, replay-based \cite{rusu2016progressive,lomonaco2017core50}, and regularization strategies \cite{liu2024fast,liu2024towards,cui2023lifelong,10825244}. Despite their effectiveness in mitigating catastrophic forgetting, these approaches still face fundamental limitations. Primarily, previous methods rely heavily on human-designed heuristics to estimate the importance of nodes and relations when transferring knowledge across evolving graph snapshots. Such handcrafted weighting schemes often do not align accurately with the true optimization objective, leading to suboptimal preservation and adaptation of knowledge between snapshots. Moreover, these methods typically require extensive computational resources due to explicit graph traversals or iterative importance computations for each node and relation. Consequently, they suffer from slow training times and substantial memory usage, rendering them inefficient and difficult to scale for large-scale KGs.

To address these limitations, we propose a novel \textbf{E}fficient \textbf{Task}-driven \textbf{T}okens for \textbf{C}ontinual \textbf{K}nowledge \textbf{G}raph \textbf{E}mbedding (ETT-CKGE). Rather than relying on predefined node/relation importance ranking rules, we introduce task-driven tokens that learn to assess the importance of nodes and relations directly from the task loss. These tokens interact with the graph embeddings and subsequently generate a token-masked embedding, which is optimized during training. These learned tokens inherently capture the task-relevant components of the graph and produce an importance mask that can be seamlessly transferred to guide learning in future snapshots. This approach offers two key advantages: it aligns importance estimation with task objectives instead of human-designed importance heuristics, and it significantly reduces training time and memory usage by replacing the graph traversal with a single matrix multiplication. 

\begin{table}[h]
\centering
\caption{Comparison of regularization-based methods for graph search space}
\label{tab:comparison}
\begin{tabular}{lcc}
\toprule
Method & Graph Traversal & Weighting Metrics \\ 
\midrule
LKGE  & Full  & Frequency  \\
FMR   & Full  & Frequency \& Gradient \\
IncDE & Partial  & Centrality    \\
FastKGE  & Partial  & Rank of centrality \\
\textbf{ETT-CKGE (Ours)} & \textbf{None}  & \textbf{Task-driven} \\ 
\bottomrule
\end{tabular}
\end{table}

\begin{figure}[t]
    \centering
    \includegraphics[width=0.8\columnwidth]{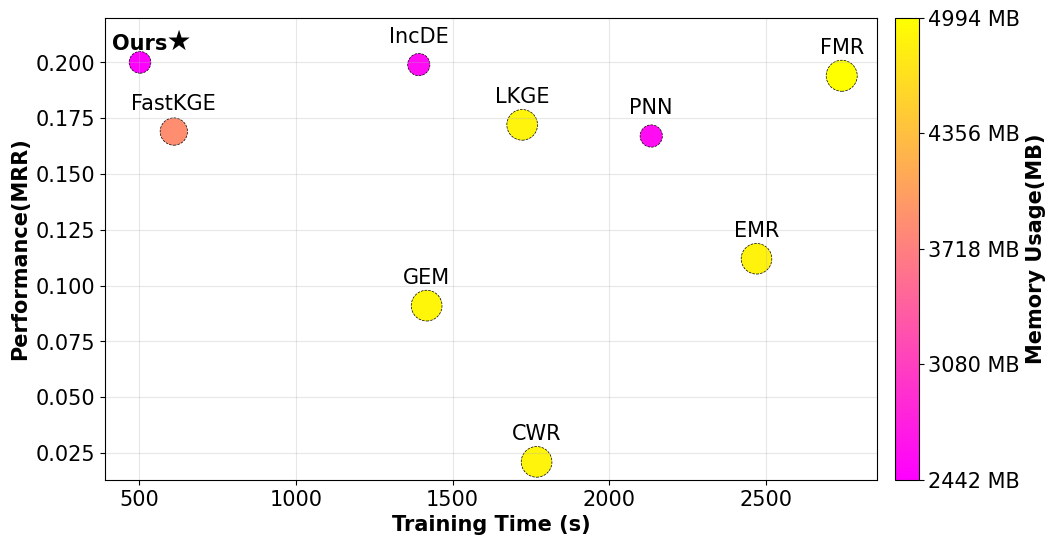}
\caption{(\textbf{Best view in color}) Comparison of performance(MRR), training time(S), and memory usage(MB) across CKGE methods on the RELATION data set. Our method achieves the best balance, delivering high accuracy with significantly reduced training time and memory consumption. The color scale indicates memory usage, with darker colors representing lower memory overhead.}
    \label{fig:tt_mrr}
\end{figure}

Moreover, Table~\ref{tab:comparison} summarizes a comparison of advanced regularization-based methods. In particular, prior methods require either full or partial graph traversal and depend on handcrafted metrics to guide learning. In contrast, our proposed method removes the need for explicit graph traversal and human-designed metrics by introducing task-driven tokens. These tokens learn to identify critical entities and relations based solely on task loss, producing a token-masked embedding that adaptively highlights relevant knowledge.

As illustrated in Figure~\ref{fig:tt_mrr}, our method consistently outperforms CKGE baselines while requiring significantly less training time and memory usage. This improvement reflects not only computational efficiency but also enhanced scalability and smoother adaptation to evolving knowledge graphs, making our approach more practical for real-world, large-scale continual learning scenarios. The main contributions of this work are summarized as follows:

\begin{itemize}
    \item We introduce a novel task-driven token module that learns to estimate the importance of nodes and relations directly from task loss. These task-guided tokens are then used to generate importance masks, enabling an effective knowledge transfer method that selectively preserves and adapts task-relevant information across growing KG snapshots without relying on human-crafted heuristics or static graph metrics.
    \item ETT-CKGE eliminates the need for graph traversal or iterative importance scoring by formulating importance estimation as a single matrix multiplication. This design significantly reduces computational overhead, achieves better scalability, and enables seamless integration with large-scale KGs, offering a practical and resource-efficient solution for CKGE settings.
    \item We conduct comprehensive experiments on six datasets with different data distributions, showing that ETT-CKGE consistently achieves competitive or superior performance in predictive accuracy while reducing training time and memory consumption compared to SOTA methods.
\end{itemize}

\section{Related Work}

Unlike standard KGE methods~\cite{bordes2013translating,trouillon2016complex,kazemi2018simple,wang2019logic}, which assume a static graph structure, CKGE is designed for dynamically evolving KGs. A recent survey~\cite{zhang2024continual} categorizes CKGE methods into three main strategies: parameter isolation methods, replay-based methods, and regularization-based methods.  

Firstly, replay-based methods~\cite{lopez2017gradient,zhou2021overcoming} replay past graph snapshots to retain information while learning new facts. However, these methods suffer from scalability issues as the memory required to store past knowledge increases significantly over time, making them impractical for large-scale applications. Secondly, parameter isolation methods, such as progressive neural networks (PNNs)~\cite{rusu2016progressive} and dynamically expandable networks (DEN)~\cite{yoon2017lifelong}, allocate separate parameter subsets to different tasks to prevent interference. While effective in avoiding catastrophic forgetting, these models require continuous expansion, leading to uncontrolled growth in model size. Lastly, regularization-based methods address catastrophic forgetting by constraining updates to critical parameters. Early approaches, such as elastic weight consolidation (EWC)~\cite{kirkpatrick2017overcoming}, used parameter importance-based constraints, while R-EWC~\cite{liu2018rotate} improved knowledge consolidation through rotation-based constraints. More recent methods, such as FMR~\cite{10825244}, leverage rotational techniques to enhance stability in CKGE, and IncDE~\cite{liu2024towards} explicitly preserve graph structure to improve retention. Moreover, FastKGE~\cite{liu2024fast} introduced low-rank adapters (LoRA) to CKGE, enabling efficient adaptation to new knowledge while reducing training time. However, FastKGE relies heavily on degree centrality within layers, requiring substantial memory to store layer information. 

As shown in Table~\ref{tab:comparison}, other SOTA regularization-based methods, such as LKGE, FMR, IncDE, and FastKGE, also depend on graph traversal—some require full-graph traversal, while others operate on partitioned graphs. This reliance introduces considerable computational costs, particularly as the KG size increases. Unlike previous methods that rely on heuristic metrics to measure the informative knowledge to overcome the forgetting issues in CKGE, we propose a set of efficient and task-driven tokens to adaptively locate essential graph components without requiring exhaustive graph traversal. By leveraging pre-trained tokens to capture global knowledge with minimal overhead, our method achieves significantly better efficiency and scalability.





\section{Continual Knowledge Graph Embedding}

\paragraph{\textbf{Problem Definition}:} A growing knowledge graph is modeled as a sequence of snapshots: $\mathcal{G} = \{\mathcal{G}_1, \mathcal{G}_2, \dots, \mathcal{G}_I\}$, where $I$ is the total number of snapshots. Each snapshot $\mathcal{G}_i$ represents a static KG at time step $i$, defined as $\mathcal{G}_i = \{\mathcal{T}_i, \mathcal{E}_i, \mathcal{R}_i\}$, where $\mathcal{T}_i$, $\mathcal{E}_i$, and $\mathcal{R}_i$ denote the sets of triplets, entities, and relations, respectively. In this context, entities represent the nodes of the graph, while relations define the semantic edges that connect them. The numbers of entities and relations in each snapshot are denoted as $N_E$ and $N_R$, respectively. A triplet $(h, r, t)$ consists of a head entity $h$, a relation $r$, and a tail entity $t$, forming a directed semantic connection. The set of triplets in snapshot $\mathcal{G}_i$ is given by $\mathcal{T}_i = \{(h, r, t) \mid (h, r, t) \in \mathcal{E}_i \times \mathcal{R}_i \times \mathcal{E}_i\}$. Each snapshot $\mathcal{G}_i$ introduces incremental knowledge in the form of newly added entities, relations, and triplets compared to the previous snapshot $\mathcal{G}_{i-1}$.


\paragraph{\textbf{Inference}:} The primary task in this work is link prediction. We assess the model's accuracy within the dynamic context of evolving KGs. Specifically, for each test fact \((h, r, t)\) in a given snapshot \(\mathcal{G}_i\), we construct two types of queries: \((h, r, \textbf{?})\) and \((\textbf{?}, r, t)\).

\section{Our Approach: Efficient Task-diven Tokens}
\begin{figure*}[t]
  \centering
    \includegraphics[width=0.99\textwidth]{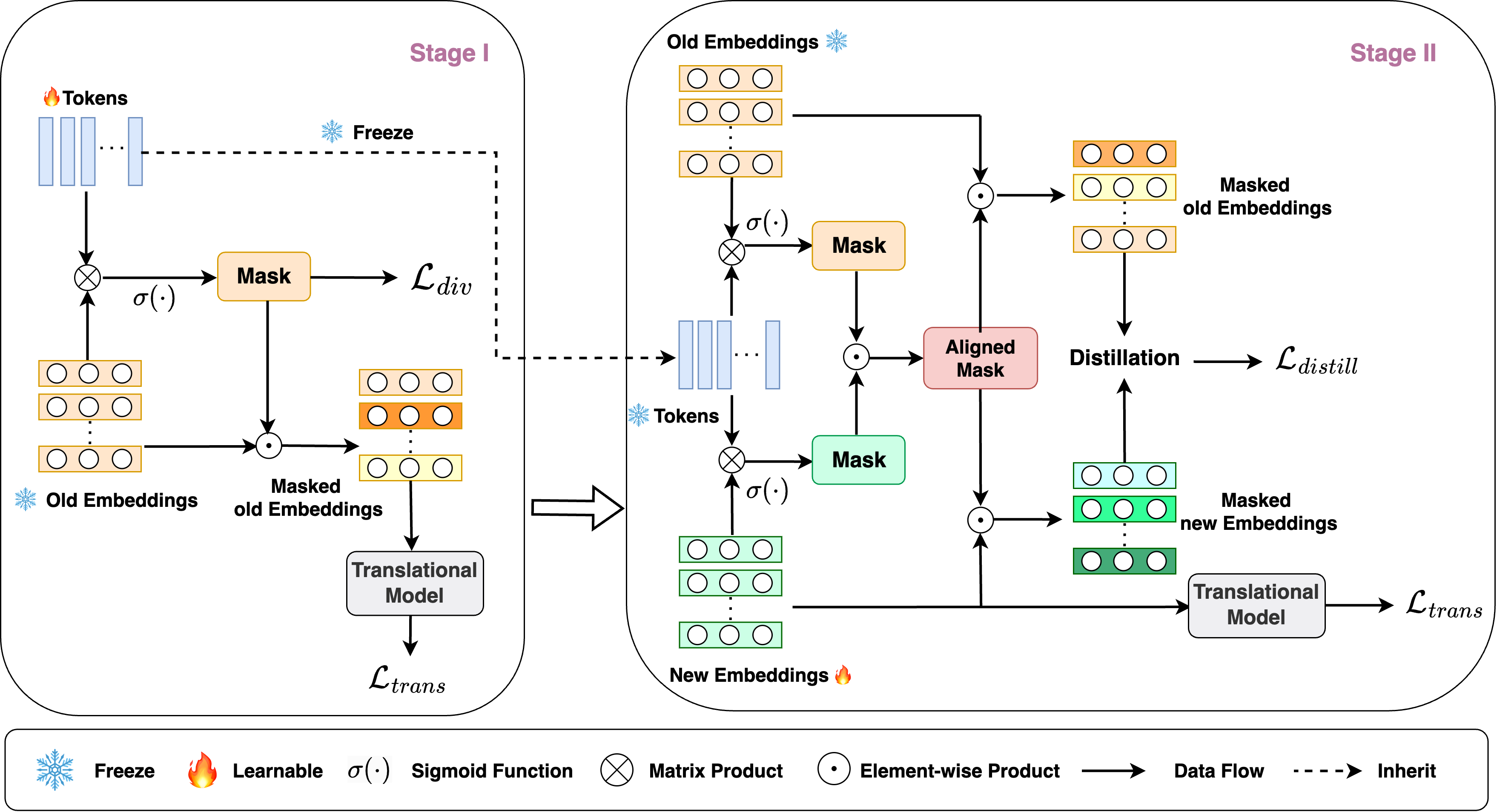}
  \caption{An overview of the ETT-CKGE framework. Stage I focuses on token pre-training, where tokens interact with previous embeddings to capture and retain prior knowledge. In Stage II, the learned tokens mask both old and new knowledge, aligning them to facilitate knowledge distillation and ensure effective continual learning}
  \label{fig:overview}
\end{figure*}
The overall architecture of ETT-CKGE is illustrated in Figure~\ref{fig:overview}, consisting of two stages: Stage I for token learning and Stage II for knowledge transfer, which are detailed in Sections~\ref{task_learning} and~\ref{mask_distillation}, respectively. 

In Stage I, we introduce a set of learnable tokens that act as task-driven representations, interacting with previously learned embeddings to capture task-relevant knowledge. Stage II focuses on continual knowledge transfer. In this stage, the previously learned embeddings and tokens are frozen to preserve historical knowledge, while only the embeddings of the new snapshot are updated. The learned tokens serve as task-driven guidance signals, promoting consistency across evolving graph snapshots. A distillation loss is applied to guide the transfer process and mitigate catastrophic forgetting. Importantly, our method achieves high computational efficiency by eliminating the need for graph traversal or iterative importance scoring, knowledge transfer is achieved through simple matrix multiplication and element-wise operations, enabling fast and scalable adaptation. Although we present the two stages separately for clarity, the token learning stage incurs negligible overhead, and the entire process can be seamlessly implemented as a single unified training pipeline in practice.

\subsection{Task-driven Token Learning}
\label{task_learning}
At snapshot $i$, we aim to extract and preserve critical structural knowledge from previously learned embeddings to guide future learning. Let the embedding matrix of entities or relations from snapshot $i-1$ be denoted as $\mathbf{E}_{i-1} \in \mathbb{R}^{N \times D}$, where $N$ corresponds to the number of entities ($N_E$) or relations ($N_R$), and $D$ is the embedding dimension.

To capture most task-relevant knowledge without relying on heuristic importance metrics in prior works~\cite{cui2023lifelong,liu2024towards}, we introduce a set of learnable tokens $\mathbf{Z} \in \mathbb{R}^{T \times D}$, guided by the task objective, where $T$ denotes the number of tokens. These tokens serve as trainable attention mechanisms that interact with the old embeddings to identify salient components in the graph. The interaction between the tokens and the old embeddings is computed via inner product in the latent space, followed by a sigmoid activation which produces a soft importance mask. For matrix multiplication compatibility, we first transpose $\mathbf{E}_{i-1}$ resulting in:

\begin{equation} 
    \mathbf{M}_{i-1} = \sigma(\mathbf{Z} \mathbf{E}_{i-1}), 
\end{equation}

\noindent where $\mathbf{M}_{i-1} \in \mathbb{R}^{T \times N}$ represents the soft mask matrix indicating the importance of each entity (or relation) with respect to each token. $\sigma(\cdot)$ is the element-wise sigmoid function. Our approach achieves importance estimation through a single matrix multiplication, offering significantly improved computational efficiency and scalability. To propagate the mask signal into the learning process, we generate masked embeddings $\mathbf{\hat{E}}_{i-1}$ by applying a token-wise weighted sum over the original embeddings $\mathbf{E}_{i-1}$:

\begin{equation} 
\mathbf{\hat{E}}_{i-1} = \sum_{t=1}^{T} \mathbf{M}_{i-1,t} \odot \mathbf{E}_{i-1}, 
\end{equation}

\noindent where $t$ indexes the $T$ token masks. The resulting $\hat{\mathbf{E}}_{i-1} \in \mathbb{R}^{N \times D}$ serves as the token-guided version of the embedding to replace $\mathbf{E}_{i-1}\in \mathbb{R}^{N \times D}$ for downstream optimization. The quality of the learned mask is implicitly guided by the translational loss $\mathcal{L}_{trans}$, which measures the effectiveness of the masked embeddings in modeling the knowledge graph structure. By freezing and optimizing only the task-driven tokens $Z$, we encourage these tokens to emphasize the most informative elements in $\mathbf{E}_{i-1}$ for task success. However, without explicit regularization, all tokens may converge to attend to the same substructures, resulting in redundant guidance. To address this, we introduce a diversity-promoting regularization based on the Dice coefficient which encourages the tokens to specialize in different graph components. Given two mask vectors $\mathbf{M}_{j} \in \mathbb{R}^{N}$ and $\mathbf{M}_{k} \in \mathbb{R}^{N}$, the diversity loss is defined as:

\begin{equation}
    \mathcal{L}_{\text{div}} = \frac{1}{T(T - 1)} \sum_{j=1}^{T} \sum_{k=1}^{T}
    \frac{2 \sum_{n=1}^{N} \mathbf{M}_{j,n} \mathbf{M}_{k,n}}{\sum_{n=1}^{N} \mathbf{M}_{j,n}^2 + \sum_{n=1}^{N} \mathbf{M}_{k,n}^2},
\end{equation}

\noindent where $j \ne k$. The diversity loss $\mathcal{L}_{\text{div}} \in [0, 1]$ penalizes high similarity between different tokens, encouraging each token to focus on distinct graph structures. By minimizing $\mathcal{L}_{\text{div}}$, we aim for the learned masks to distribute their attention across different substructures in the KG. This formulation penalizes high overlap between any two token masks, thereby enforcing diversity in their attention distributions. 

To align token training with the predictive task, we adopt the TransE~\cite{bordes2013translating} as a translational model to learn the KGs in the current snapshot. The original TransE loss is defined as:
\begin{equation}
    \mathcal{L}_{trans} = \sum_{(h, r, t) \in \mathcal{T}} \sum_{(h', r, t') \in \mathcal{T'}} \max( 0,\gamma + f(\mathbf{h},\mathbf{r}, \mathbf{t})
    - f(\mathbf{h'}, \mathbf{r}, \mathbf{t'})),
    \label{margin_rank_loss}
\end{equation}

\noindent where $\mathcal{T}$ represents the set of positive triplets, and $\mathcal{T'}$ denotes the set of negative triplets generated for negative sampling. The parameter $\gamma$ is a margin hyperparameter that controls the separation between positive and negative triplets. The score function is defined as: $f(\mathbf{h},\mathbf{r},\mathbf{t}) = ||\mathbf{h}+\mathbf{r}-\mathbf{t}||_2^2$. 
The final objective for token learning in Stage I is defined as:

\begin{equation} \mathcal{L}_{\text{token}} = \mathcal{L}_{\text{trans}} + \lambda \mathcal{L}_{\text{div}}, 
\end{equation}

\noindent where $\lambda$ is a hyperparameter balancing the loss terms. This formulation enables tokens to capture both task-relevant and diverse structural patterns, serving as a lightweight mechanism to guide continual learning without expensive traversal or handcrafted rules.

\subsection{Distillation via Learned Token Masks}
\label{mask_distillation}
Building on the effectiveness of the learned task-driven tokens in identifying key graph substructures, we further leverage these tokens to facilitate efficient and targeted knowledge distillation across growing knowledge graph snapshots. In this stage, we freeze the learned task-driven tokens $\mathbf{Z}$ and the old embeddings $\mathbf{E}_{i-1}$, thus preserving the saliency patterns learned from prior snapshots. Only the new embeddings $\mathbf{E}_{i}$ are updated during training, allowing it to align with task-relevant structural components identified by the learned tokens. To capture informative knowledge from both the old and new snapshots, we compute respective token-guided importance masks as: 

\begin{equation} 
\mathbf{M}_{i-1} = \sigma(\mathbf{Z} \mathbf{E}_{i-1}), \quad \mathbf{M}_i = \sigma(\mathbf{Z} \mathbf{E}_i), 
\end{equation} 

\noindent where $\mathbf{M}_{i-1}, \mathbf{M}_i \in \mathbb{R}^{T \times N}$ represent the attention masks derived from previous and current embeddings, respectively. It is important to note that knowledge transfer is applied only to those transposed embeddings in $\mathbf{E}_i$ corresponding to entities and relations that also existed in the previous snapshot $\mathbf{E}_{i-1}$. In contrast, new entities and relations introduced in snapshot $i$ are learned purely through the task loss $\mathcal{L}_{\text{trans}}$, without any distillation guidance. This design ensures that distillation focuses solely on preserving previously acquired knowledge, while allowing the model to flexibly accommodate new information.

\paragraph{\textbf{Aligned token masks.}} Direct application of these independent masks may lead to structural misalignment, where salient components differ across snapshots. To address this, we introduce a token-level alignment mechanism, forming a joint mask $\mathbf{M} = \mathbf{M}_{i-1} \odot \mathbf{M}_i$. This aligned mask is applied to both $\mathbf{E}_{i-1}$ and $\mathbf{E}_{i}$ to emphasize consistently critical entity and relation embeddings across snapshots, ensuring that distillation focuses on components deemed important by both the previous and current knowledge, rather than on noisy or transient elements. Based on this, we formulate the knowledge distillation as:

\begin{equation}
\mathcal{L}_{distill} = \frac{1}{TN} \sum_{t=1}^{T} \mathbf{M}_{t} \odot \Big\| \mathbf{E}_{i-1} - \mathbf{E}_i \Big\|^2_2,
\end{equation}

\noindent where the L2 norm quantifies the divergence between matched embeddings, and the aligned mask $\mathbf{M}$ selectively emphasizes structurally important graph components during distillation. In contrast to prior methods that require explicit graph traversal or iterative computations over all triples to estimate importance, our approach performs this step through a single matrix operation, yielding substantial improvements in computational efficiency.

\paragraph{\textbf{Overall loss function.}} The overall training loss in this stage combines the task-specific translational loss with the distillation loss:
\begin{equation} 
    \mathcal{L} = \mathcal{L}_{\text{trans}} + \alpha \mathcal{L}_{\text{distill}}, 
    \end{equation} 

\noindent where $\alpha$ is a hyperparameter controlling the strength of the distillation.

\section{Experiment and Analysis}

\begin{table*}[t]
\centering
\footnotesize 
\setlength{\tabcolsep}{3pt} 
\caption{The statistics of datasets.}
\label{tab:dataset_statistics}
\resizebox{\textwidth}{!}{%
\begin{tabular}{lccccccccccccccc}
\toprule
\multirow{2}{*}{Dataset} & \multicolumn{3}{c}{Snapshot 0} & \multicolumn{3}{c}{Snapshot 1} & \multicolumn{3}{c}{Snapshot 2} & \multicolumn{3}{c}{Snapshot 3} & \multicolumn{3}{c}{Snapshot 4} \\
\cmidrule(lr){2-4} \cmidrule(lr){5-7} \cmidrule(lr){8-10} \cmidrule(lr){11-13} \cmidrule(lr){14-16}
& $N_E$  & $N_R$  & $N_T$  & $N_E$  & $N_R$  & $N_T$  & $N_E$  & $N_R$  & $N_T$  & $N_E$  & $N_R$  & $N_T$  & $N_E$  & $N_R$  & $N_T$  \\
\midrule
ENTITY   &  2909  &  233  &  46388  &  5817  &  236  &  72111  &  8275  &  236  &  73785  & 11633  &  237  &  70506  & 14541  &  237  &  47326  \\
RELATION & 11560  &   48  &  98819  & 13343  &   96  &  93535  & 13754  &  143  &  66136  & 14387  &  190  &  30032  & 14541  &  237  &  21594  \\
FACT     & 10513  &  237  &  62024  & 12779  &  237  &  62023  & 13586  &  237  &  62023  & 13894  &  237  &  62023  & 14541  &  237  &  62023  \\
HYBRID   &  8628  &   86  &  57561  & 10040  &  102  &  20873  & 12779  &  151  &  88017  & 14393  &  209  & 103339  & 14541  &  237  &  40326  \\
FB-CKGE  &  7505  &  237  & 186070  & 11258  &  237  &  31012  & 13134  &  237  &  31012  & 14072  &  237  &  31012  & 14541  &  237  &  31010  \\
WN-CKGE  & 24567  &   11  &  55801  & 28660  &   11  &   9300  & 32754  &   11  &   9300  & 36848  &   11  &   9300  & 40943  &   11  &   9302  \\
\bottomrule
\end{tabular}
}
\raggedright{\tiny{$N_E$, $N_R$ and $N_T$ denote the number of cumulative entities and relations, and current triples in each snapshot $i$.}}
\end{table*}

\subsection{Datasets}

We evaluate the proposed ETT-CKGE framework on six benchmark datasets: ENTITY, RELATION, FACT, HYBRID, FB-CKGE, and WN-CKGE. The first four datasets, introduced in~\cite{cui2023lifelong}, represent different types of knowledge growth in CKGE: ENTITY tracks increasing entities, RELATION focuses on evolving relations, FACT captures growing knowledge triples, and HYBRID combines all three. FB-CKGE and WN-CKGE were introduced by~\cite{liu2024fast} as continual extensions of FB15K and WN18~\cite{bordes2013translating}. We set the number of snapshots for all datasets to 5, with the train/validation/test split ratio fixed at 3:1:1. Dataset statistics are provided in Table~\ref{tab:dataset_statistics}.

\subsection{Baselines}

We compare ETT-CKGE with a range of continual learning baselines, including \textbf{fine-tune}, \textbf{parameter-isolation}, \textbf{replay-based}, and \textbf{regularization-based} methods. Notably, the Fine-Tune baseline simply continues training the KGE model on new incoming data without any explicit mechanism to preserve previously learned knowledge. As a result, it is efficient in terms of training time but suffers from severe forgetting. The remaining baselines implement different strategies to mitigate catastrophic forgetting and preserve prior knowledge. Together, they provide a comprehensive framework to evaluate the effectiveness and efficiency of ETT-CKGE in continual knowledge graph embedding.



\subsection{Experimental Setup}

All experiments were conducted using PyTorch on a single NVIDIA A6000 GPU. Experiments were conducted using a batch size selected from \{1024, 2048, 3072\}, and a learning rate chosen from \{0.01, 0.001, 0.0001, 0.00001\}. The Adam optimizer is used for all experiments. The hyperparameter $\alpha$ varies across different datasets, ranging from 1,000 to 100,000, while $\lambda$ is selected from the range [0, 1]. The margin $\gamma$ is set to 9, and $D$ for all experiments is set to 200. In all experiments, the token number $T$ is set to different integer values within the range (0,10]. For fairness, we run all baseline models on each benchmark dataset five times to take their average performance and fine-tune their hyperparameters to report the best performance. The code and hyperparameter settings are available at 
\href{https://github.com/lijingzhu1/ETT-CKGE}{Github}.


\subsection{Evaluation Metrics}

We evaluate ETT-CKGE using three metrics: \textbf{Mean Reciprocal Rank (MRR)}, \textbf{Hits@$k$}, and \textbf{Training Time}. MRR measures the average inverse rank of the correct entity, while Hits@$k$ indicates the proportion of correct entities ranked in the top $k$ predictions. Training Time reflects the total time required to train the model across all snapshots. We report MRR, Hits@$k$ (with $k \in \{1, 10\}$), and training time to assess both performance and efficiency.
\begin{table*}[t]
\centering
\caption{Main experimental results}
\label{tab:results_fasterkge_combined}
\resizebox{\textwidth}{!}{%
\begin{tabular}{lcccccccccccc}
\toprule
Model & \multicolumn{4}{c}{ENTITY} & \multicolumn{4}{c}{RELATION} & \multicolumn{4}{c}{FACT} \\
\cmidrule(lr){2-5} \cmidrule(lr){6-9} \cmidrule(lr){10-13}
 & MRR & H@1 & H@10 & Training Time (s) 
 & MRR & H@1 & H@10 & Training Time (s)
 & MRR & H@1 & H@10 & Training Time (s) \\
\midrule
Fine-Tune & 0.171 & 0.093 & 0.319 & 464 & 0.085 & 0.036 & 0.170 & 419 & 0.169 & 0.092 & 0.323 & 305 \\
\midrule
PNN~\cite{rusu2016progressive}& 0.229 & 0.130 & 0.425 & 2145 & 0.167 & 0.096 & 0.305 & 2134 & 0.157 & 0.084 & 0.290 & 1613 \\
CWR~\cite{lomonaco2017core50} & 0.087 & 0.028 & 0.200 & 2350 & 0.021 & 0.010 & 0.043 & 1768 & 0.082 & 0.028 & 0.194 & 2753 \\
\midrule
GEM~\cite{lopez2017gradient} & 0.165 & 0.085 & 0.321 & 1993 & 0.091 & 0.039 & 0.191 & 1417 & 0.174 & 0.091 & 0.344 & 1139 \\
EMR~\cite{wang2019sentence} & 0.173 & 0.065 & 0.333 & 4177 & 0.112 & 0.053 & 0.226 & 2740 & 0.170 & 0.090 & 0.335 & 1722 \\
\midrule
LKGE~\cite{cui2023lifelong} & 0.240 & 0.141 & 0.434 & 2374 & 0.172 & 0.093 & 0.343 & 1722 & 0.210 & 0.122 & 0.389 & 1090 \\
FMR~\cite{10825244} & 0.253 & 0.138 & 0.450 & 3094 & 0.194 & 0.107 & 0.367 & 2742 & 0.215 & 0.128 & 0.392 & 1661 \\
IncDE~\cite{liu2024towards} & \underline{0.253} & \underline{0.151} & \underline{0.448} & 1587 & \underline{0.199} & \underline{0.110} & \underline{0.368} & 1392 & \underline{0.216} & \underline{0.128} & \underline{0.391} & 1752 \\
FastKGE~\cite{liu2024fast} & 0.230 & 0.140 & 0.404 & \underline{821} & 0.169 & 0.101 & 0.296 & \underline{610} & 0.171 & 0.105 & 0.291 & \underline{583} \\
\textbf{ETT-CKGE} & \textbf{0.260} & \textbf{0.158} & \textbf{0.456} & \textbf{784} & \textbf{0.200} & \textbf{0.112} & \textbf{0.369} & \textbf{502} & \textbf{0.217} & \textbf{0.129} & \textbf{0.396} & \textbf{506} \\
\bottomrule
\end{tabular}%
}
\vspace{10pt} 

\resizebox{\textwidth}{!}{%
\begin{tabular}{lcccccccccccc}

\toprule
Model & \multicolumn{4}{c}{HYBRID} & \multicolumn{4}{c}{FB-CKGE} & \multicolumn{4}{c}{WN-CKGE} \\
\cmidrule(lr){2-5} \cmidrule(lr){6-9} \cmidrule(lr){10-13}
 & MRR & H@1 & H@10 & Training Time (s) 
 & MRR & H@1 & H@10 & Training Time (s) 
 & MRR & H@1 & H@10 & Training Time (s) \\
\midrule
Fine-Tune & 0.137 & 0.074 & 0.256 & 559 & 0.182 & 0.098 & 0.344 & 277 &0.1 &0.004 & 0.259&392 \\
\midrule
PNN~\cite{rusu2016progressive} & 0.185 & 0.101 & 0.350 & 2039 & 0.215 & 0.122 & 0.402 & 1351 & 0.133 & 0.002 & 0.343 & 1429 \\
CWR~\cite{lomonaco2017core50} & 0.037 & 0.015 & 0.078 & 1986 & 0.072 & 0.011 & 0.187 & 2039 & 0.005 & 0.000 & 0.012 & 1265 \\
\midrule
GEM~\cite{lopez2017gradient} & 0.135 & 0.070 & 0.261 & 1804 & 0.183 & 0.098 & 0.352 & 1069 & 0.114 & 0.001 & 0.290 & 1049 \\
EMR~\cite{wang2019sentence} & 0.140 & 0.074 & 0.268 & 3154 & 0.181 & 0.097 & 0.347 & 1474 & 0.114 & 0.002 & 0.287 & 1160 \\
\midrule
LKGE~\cite{cui2023lifelong} & 0.179 & 0.111 & 0.372 & 1612 & 0.220 & 0.125 & 0.412 & 1197 & 0.139 & 0.070 & 0.333 & 1136 \\
FMR~\cite{10825244} & 0.206 & 0.121 & 0.375 & 3258 & 0.220 & 0.125 & 0.413 & 2086 & 0.132 & 0.003 & 0.324 & 1850 \\
IncDE~\cite{liu2024towards} & \underline{0.223} & \underline{0.130} & \underline{0.401} & 1675 & \underline{0.232} & \underline{0.133} & \underline{0.425} & 1447 & 0.150 & 0.004 & 0.362 & 1087 \\
FastKGE~\cite{liu2024fast} & 0.198 & 0.120 & 0.345 & \underline{841} & 0.220 & 0.128 & 0.400 & \textbf{390} & \textbf{0.160} & \textbf{0.011} & \underline{0.368} & \underline{448} \\
\textbf{ETT-CKGE} & \textbf{0.224} & \textbf{0.131} & \textbf{0.402} & \textbf{535} & \textbf{0.236} & \textbf{0.137} & \textbf{0.428} & \underline{413} & \underline{0.153} & \underline{0.080} & \textbf{0.385} & \textbf{369} \\
\bottomrule
\end{tabular}%
}
\end{table*}

Beyond performance, we assess efficiency and scalability using three metrics: \textbf{Cumulative Training time}, which reflects knowledge adaptation smoothness; \textbf{Peak Allocated memory}, which measures the maximum memory usage per snapshot; and \textbf{Updated Parameters}, which indicates the number of parameters updated during training and serves as an indicator of computational cost.

\subsection{Experimental Results \& Discussion}

Table~\ref{tab:results_fasterkge_combined} presents the experimental results across six benchmark datasets. The results demonstrate that ETT-CKGE achieves competitive performance relative to state-of-the-art continual KGE methods, without overclaiming superiority. Compared to the Fine-Tune model, ETT-CKGE yields MRR improvements ranging from 30.1\% to 135.3\%, highlighting the severity of knowledge degradation in Fine-Tune as new snapshots are introduced. Notably, despite employing a more sophisticated architecture, ETT-CKGE achieves faster training time than Fine-Tune on the HYBRID dataset. This efficiency stems from the task-driven token design, which allows our model to selectively encode and transfer essential knowledge without relying on time-consuming graph traversal, thereby reducing computational overhead while maintaining strong performance.

Compared to the second-best performing models, ETT-CKGE achieves a 50\% to 96\% reduction in training time while maintaining comparable or superior MRR. This improvement stems from ETT-CKGE’s token-guided, task-driven framework, which not only eliminates the need for expensive graph traversal and handcrafted heuristics, as seen in models like IncDE and FMR, but also enables the model to identify the most informative nodes and relations directly from task signals. This selective focus facilitates more efficient knowledge transfer and model adaptation, resulting in a highly effective and scalable continual learning approach.

\begin{figure}[!htbp]
\centering
  \includegraphics[width=0.7\textwidth]{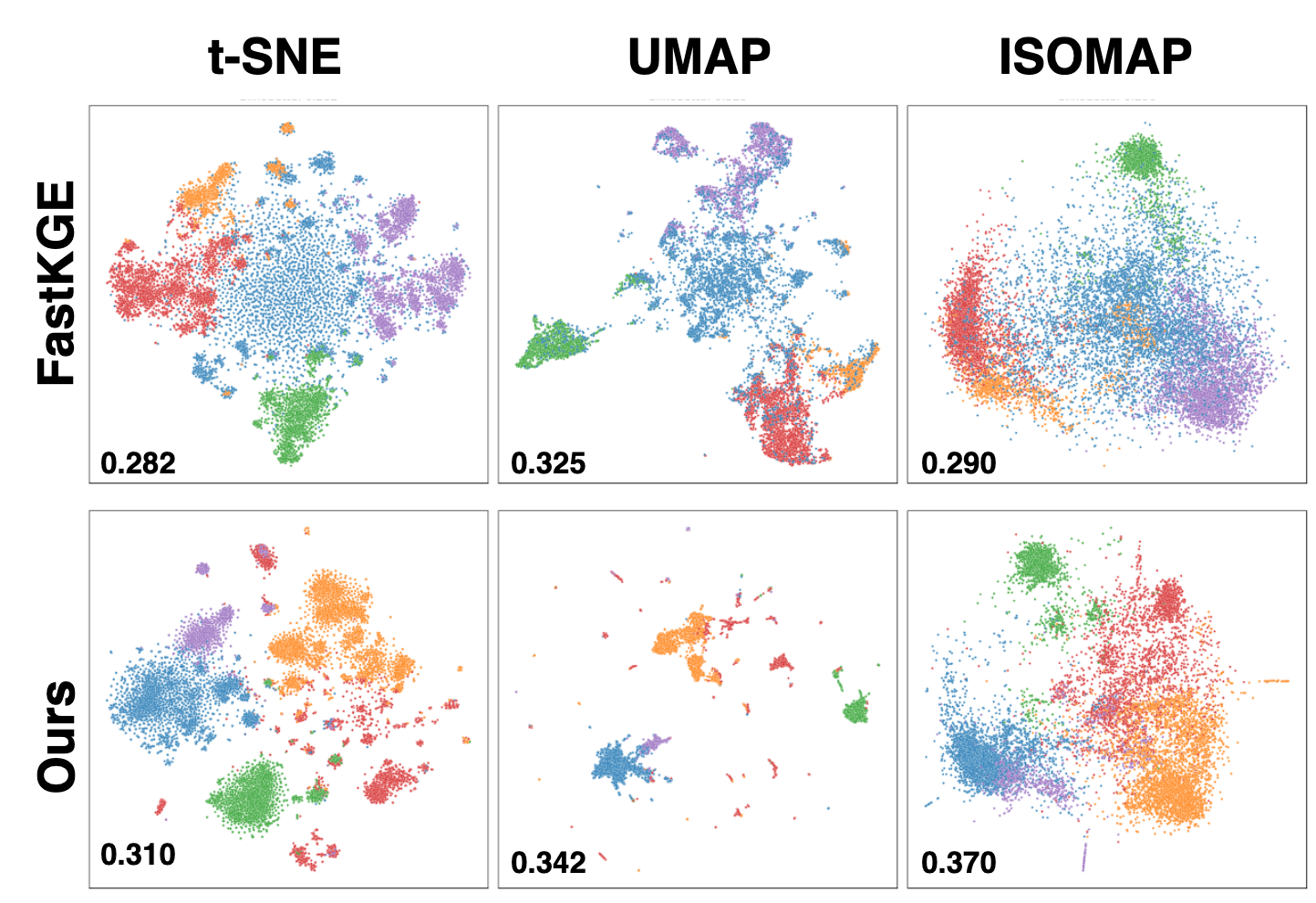}
  \caption{Entity embedding visualization on the ENTITY dataset. The Silhouette Score, shown in the lower-left corner of each plot, quantitatively reflects the clustering quality of the entity embeddings; higher scores indicate more well-separated and compact clusters.}
  \label{fig:clusters}
\end{figure}

Compared to FastKGE, the second fastest model, ETT-CKGE consistently achieves 7.2\% to 13\% higher MRR across most datasets, demonstrating its superior accuracy in CKGE. While FastKGE relies on rank-based adapters and human-designed heuristics to integrate new knowledge, its optimization is not directly aligned with the task objective. In contrast, ETT-CKGE leverages task-driven tokens that are optimized adaptively through the training loss, enabling more effective and targeted knowledge transfer. Although FastKGE performs competitively on the WN-CKGE dataset, ETT-CKGE still achieves a shorter training time, offering a better balance between performance and efficiency.

While FastKGE sacrifices model expressiveness to speed up continual learning, our method maintains both efficiency and predictive quality, making it a more well-rounded choice. Furthermore, Figure-\ref{fig:clusters} presents entity embedding visualizations via three methods, t-SNE, UMAP, and ISOMAP. It is clear that the entity embeddings learned by ETT-CKGE have more separable patterns compared to embeddings learned by FastKGE.


Generally, ETT-CKGE achieves superior or comparable performance to complex, resource-intensive SOTA models while significantly reducing training time and memory consumption, as explained in Section~\ref{efficiency_scability_analysis}. Compared to efficiency-focused approaches, ETT-CKGE demonstrates notable improvements in both accuracy and computational efficiency. Overall, ETT-CKGE offers an excellent balance between performance and efficiency, making it a practical and scalable solution for evolving knowledge graphs.



\subsection{Catastrophic Forgetting Analysis in Continual Learning}
\label{efficiency_scability_analysis}
\begin{figure}[!htbp]
\centering
  \includegraphics[width=0.9\textwidth, height =0.3 \textheight]{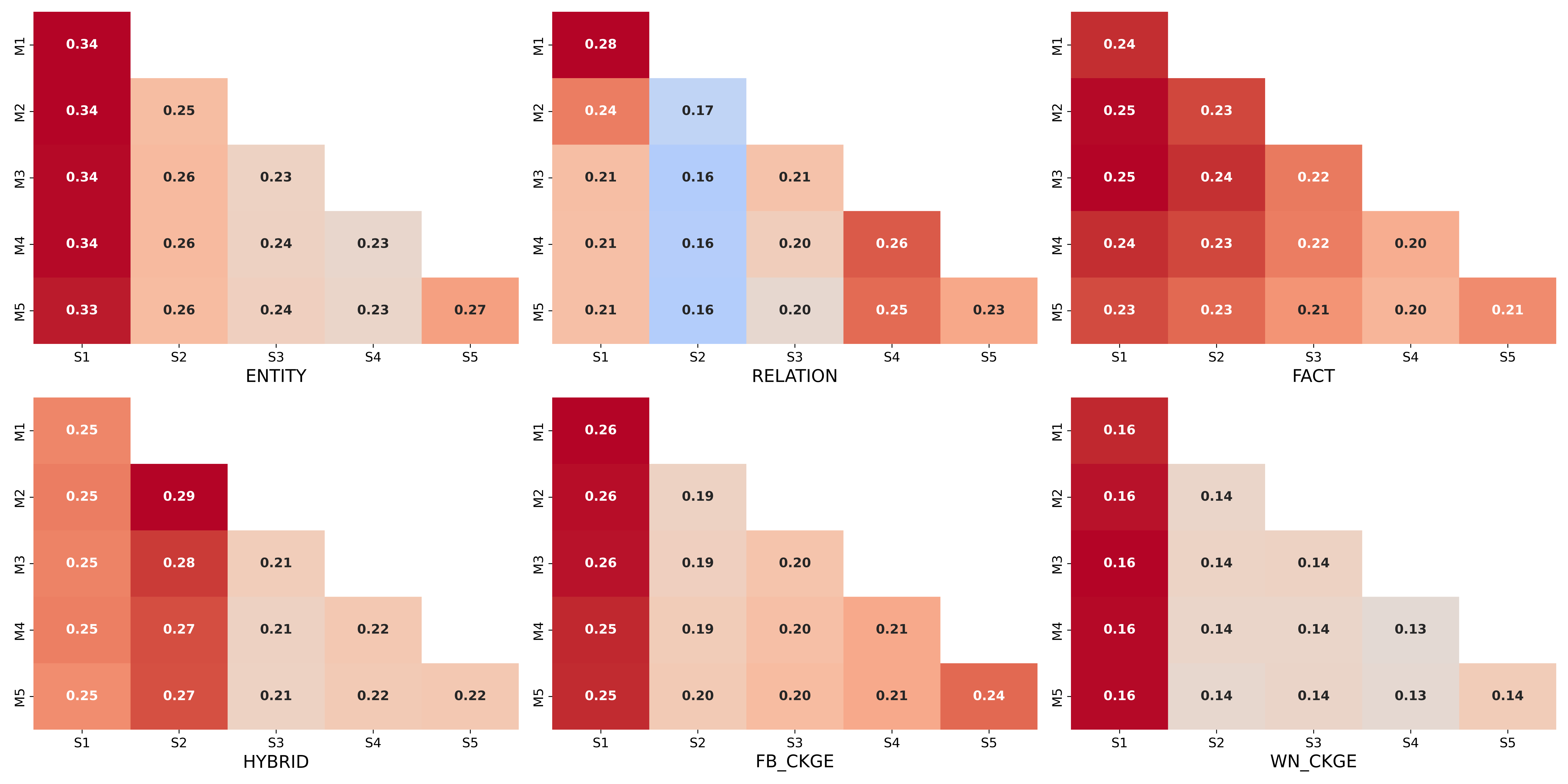}
  \caption{MRR Changes}
  \label{fig:mrr}
\end{figure}
Since catastrophic forgetting is the main concern in CKGE, Figure~\ref{fig:mrr} illustrates how our model addresses catastrophic forgetting, showcasing its performance across six datasets during sequential learning. Each heatmap displays the MRR achieved by our model on each dataset over snapshots S1 to S5. Model stages are denoted as $Mi$, where $i$ represents the snapshot number. Warmer colors indicate higher MRR, while cooler colors suggest performance degradation and forgetting on that dataset.

In summary, it demonstrates our model's effective mitigation of catastrophic forgetting via robust knowledge preservation. Although some dataset-specific MRR variations appear, especially on the relation-centric RELATION dataset, ETT-CKGE generally maintains stable MRR across sequential snapshots. Notably, on FB-CKGE and WN-CKGE datasets, ETT-CKGE exhibits remarkable resilience to forgetting, indicating a strong capability to learn and adapt to evolving KGs without significantly compromising prior knowledge. This balance of knowledge preservation and sequential learning highlights the effectiveness of ETT-CKGE in addressing catastrophic forgetting in dynamic KG scenarios.

\subsection{Efficiency and Scalability Analysis in Continual Learning}

\begin{figure*}[t]
    \centering
    \captionsetup{font=small} 

    \begin{subfigure}[b]{0.3\textwidth}
        \centering
        \includegraphics[width=\textwidth]{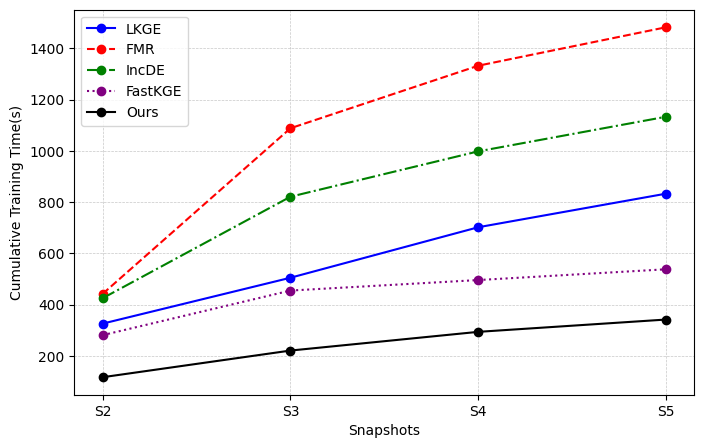}
        \caption{Cumulative Training Time}
        \label{fig:sc_a}
    \end{subfigure}
    \hfill
    \begin{subfigure}[b]{0.3\textwidth}
        \centering
        \includegraphics[width=\textwidth]{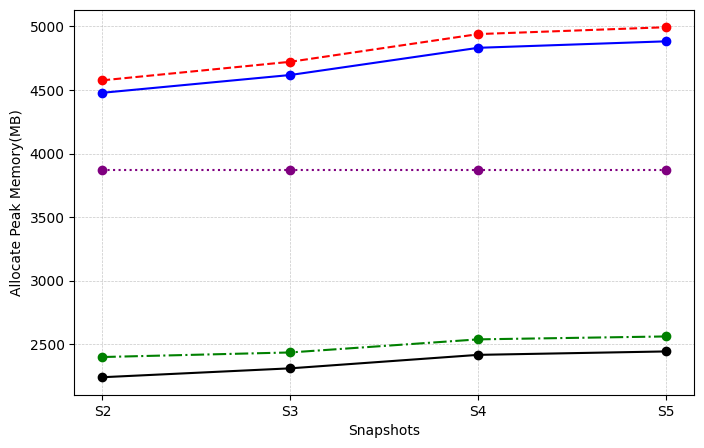}
        \caption{Peak Allocated Memory}
        \label{fig:sc_b}
    \end{subfigure}
    \hfill
    \begin{subfigure}[b]{0.3\textwidth}
        \centering
        \includegraphics[width=\textwidth]{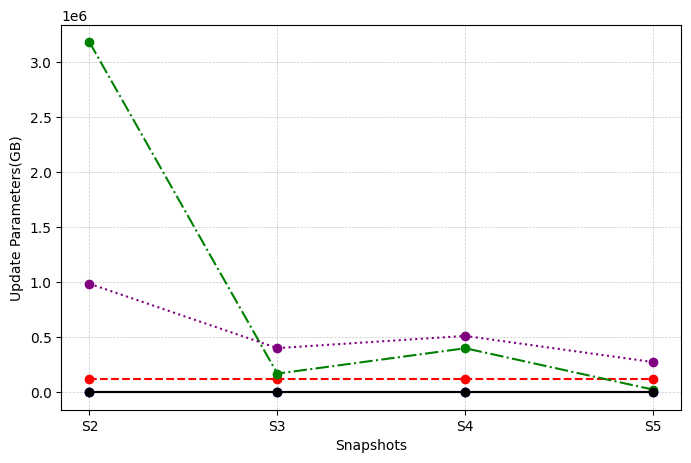}
        \caption{Update Parameters}
        \label{fig:sc_c}
    \end{subfigure}

    \caption{Model Scalability Analysis Over Time}
    \label{fig:analysis_scability}
\end{figure*}

Figure~\ref{fig:analysis_scability} provides experimental validation of the efficiency and scalability on the RELATION dataset. For a model to scale well with evolving KGs, it must adapt to new information efficiently, meaning without a large increase in computational work. We analyze efficiency and stability metrics from snapshot 2 onward to focus on model behavior in dynamic scenarios.

\begin{itemize}[label=\textbullet]
    \item \textbf{Cumulative Training Time}: As shown in Figure~\ref{fig:sc_a}, ETT-CKGE consistently achieves the lowest cumulative training time across all snapshots, outperforming even the second-fastest model, FastKGE, by 1.6 to 2.3x from S2 to S5. Beyond the raw efficiency gain, training curves show smoother adaptation to evolving knowledge graphs, with stable incremental increases in training time. This indicates that ETT-CKGE facilitates more efficient knowledge transfer between snapshots, minimizing disruption and avoiding the sharp cost spikes observed in other models. Such smooth transitions suggest that the task-driven token mechanism effectively captures and reuses informative components without requiring heavy computational overhead. While training speed is a clear advantage, this stability in knowledge adaptation underscores the broader benefit of our design, ensuring efficient, consistent, and scalable continual learning.

    \item \textbf{Peak Allocated Memory}: Figure~\ref{fig:sc_b} shows that ETT-CKGE achieves the lowest peak memory consumption among all baselines, reducing memory usage by 150MB to 2GB compared to memory-intensive models like FMR and LKGE. This efficiency comes from ETT-CKGE’s fixed-size token design and lightweight architecture, which avoids storing additional structures like entity layers or replay buffers required in other methods. 

    \textbf{Updated Parameters:} As shown in Figure~\ref{fig:sc_c}, ETT-CKGE consistently maintains a low number of updated parameters, around 2,000 across all snapshots. These parameters come from the fixed-size task-driven tokens, which are used solely to retain old knowledge and are not expanded when learning new snapshots. In contrast, models like IncDE introduce new parameters to learn additional knowledge at each snapshot, leading to significantly higher computational cost. This lightweight design allows ETT-CKGE to achieve faster training while still maintaining strong performance.
    
\end{itemize}

\subsection{Ablation Study}

Table~\ref{tab:combined_comparison} presents the ablation results of ETT-CKGE, evaluating the contribution of three key components: the distillation loss ($\mathcal{L}_{distill}$), Stage I training (SIT), and the diversity loss ($\mathcal{L}_{div}$), across various datasets. 


\textbf{Effect of Distillation Loss}: The results clearly highlight that $\mathcal{L}_{distill}$ is the core driver of ETT-CKGE’s effectiveness. Removing it leads to a significant performance drop across all datasets, underscoring its essential role in enabling task-relevant knowledge transfer. This task-driven loss directly optimizes the token-guided embedding space to capture critical information from both old and new knowledge without relying on handcrafted heuristics or traversal-based processing.

\textbf{Effect of Stage I Training}: SIT plays an important supporting role. Eliminating Stage I training slightly reduces performance, especially in FB-CKGE, while reducing training time. This shows that although SIT introduces extra computation, it strengthens the quality of token learning and improves overall performance when paired with $\mathcal{L}_{distill}$.

\textbf{Effect of Diversity Loss}: Diversity loss helps ensure varied and effective token learning. Removing $\mathcal{L}_{div}$ causes a minor drop in MRR, indicating that it contributes to performance improvements but is not as crucial as $\mathcal{L}_{dist}$. Additionally, removing $\mathcal{L}_{div}$ reduces training time, confirming that its computation introduces extra overhead.

\begin{table*}[t]
\centering
\footnotesize
\caption{Ablation results}
\resizebox{\textwidth}{!}{%
\begin{tabular}{lcccccccccccccc}
\toprule
\multirow{2}{*}{$\mathcal{L}_{distill}$} & 
\multicolumn{2}{c}{Token Training} 
& \multicolumn{2}{c}{ENTITY} & \multicolumn{2}{c}{RELATION} & \multicolumn{2}{c}{FACT} 
& \multicolumn{2}{c}{HYBRID} & \multicolumn{2}{c}{FB-CKGE} & \multicolumn{2}{c}{WN-CKGE} \\
\cmidrule(lr){2-3} \cmidrule(lr){4-5} \cmidrule(lr){6-7} \cmidrule(lr){8-9} \cmidrule(lr){10-11} \cmidrule(lr){12-13} \cmidrule(lr){14-15}
 &SIT & $\mathcal{L}_{div}$ & MRR & T(s) & MRR & T(s) & MRR & T(s) & MRR & T(s) & MRR & T(s) & MRR & T(s) \\
\midrule
\ding{51}  & \ding{51} & \ding{51}  &  0.260 & 784    & 0.200 & 502 & 0.217 & 506    & 0.222 & 559 & 0.236 & 413 & 0.153 & 369 \\
\ding{51}  & \ding{51} & \ding{55}   & 0.258 & 722    & 0.194 & 496 & 0.215 & 461    & 0.220 & 567 & 0.233 & 405 & 0.152 & 343 \\
\ding{51}  & \ding{55} & \ding{55}   & 0.257 & 612    & 0.193 & 448 & 0.215 & 335    & 0.220 & 497 & 0.221 & 316 & 0.150 & 341 \\
\ding{55}  & \ding{55} & \ding{55}   & 0.170 & 528    & 0.085 & 417 & 0.161 & 287    & 0.138 & 406 & 0.178 & 275 & 0.102 & 410 \\

\bottomrule
\end{tabular}%
}

\label{tab:combined_comparison}
\end{table*}

\section{Conclusion Remarks and Future Work}

This paper introduces a novel regularization-based CKGE model with a self-guided token mechanism for better efficiency and performance. The proposed model significantly reduces the adaptation time between snapshots and memory costs, thus opening the door to real-world applications with large data volumes. As evidenced by extensive comparative experiments and an ablation study, the proposed model outperforms the SOTA models in both predictive accuracy and model efficiency. In addition, our model can be adapted to downstream tasks at all levels, including link prediction, node classification, and graph classification. 

In the future, while improving efficiency is critical for practical applications, enhancing robustness to noise and high sparsity in graphs is another challenge to be solved. With current advances in Large Language Models (LLMs)~\cite{brown2020language,chowdhery2023palm,touvron2023llama} and Multi-Modal Learning (MML)~\cite{radford2021learning,jia2021scaling}, leveraging knowledge foundations of built LLMs via MML has become a promising approach~\cite{yasunaga2022linkbert,liu2023graphprompt} to handle noisy and sparse graphs. Moreover, CKGE plays a vital role in graph foundation models in continually evolving domains, such as recommender systems, social networks, and biomedical knowledge reasoning. 

\section{Acknowledgments}
The authors would like to thank the Department of Computer Science at Bowling Green State University, the Department of Artificial Intelligence \& Informatics at Mayo Clinic, and the Department of Computer Science at University of Houston Clear Lake, and the Department of Ctr-Secure AI for Healthcare at The University of Texas Health Science, Houston, for the support and resources that contributed to this research. This work is supported by projects, R24ES036131, R01AT012871, and U24AI171008.

\bibliographystyle{splncs04}
\bibliography{references}

\end{document}